%% file: lrec2022-example.tex
\documentclass[10pt, a4paper]{article}
\usepackage{lrec2022} 
\usepackage{multibib}
\newcites{languageresource}{Language Resources}
\usepackage{graphicx}
\usepackage{tabularx}
\usepackage{multirow}
\usepackage{subcaption}
\usepackage{soul}
\usepackage{titlesec}
\titleformat{\section}{\normalfont\large\bfseries\center}{\thesection.}{1em}{}
\titleformat{\subsection}{\normalfont\SmallTitleFont\bfseries\raggedright}{\thesubsection.}{1em}{}
\titleformat{\subsubsection}{\normalfont\normalsize\bfseries\raggedright}{\thesubsubsection.}{1em}{}
\renewcommand\thesection{\arabic{section}}
\renewcommand\thesubsection{\thesection.\arabic{subsection}}
\renewcommand\thesubsubsection{\thesubsection.\arabic{subsubsection}}

\usepackage{lucasstyle}
\usepackage{epstopdf}
\usepackage[utf8]{inputenc}

\usepackage{hyperref}
\usepackage{xstring}

\usepackage{color}

\title{Unsupervised Word Segmentation from Discrete Speech Units in Low-Resource Settings}

\name{Marcely Zanon Boito$^{1,*}$, Bolaji Yusuf$^{2,3}$, Lucas Ondel$^{4}$, \\\large\textbf{Aline Villavicencio$^5$, Laurent Besacier$^{6}$}}

\address{
  $^1$Avignon University, FR, 
  $^2$Bogazici University, TR\\ 
  $^3$Brno University of Technology, CZ \\
  $^4$LISN CNRS, FR 
  $^5$University of Sheffield, UK\\
  $^6$Naver Labs Europe, FR and University Grenoble Alpes, FR \\
  \**~Research done while at University Grenoble Alpes.\\
  \textbf{contact:} marcely.zanon-boito at univ-avignon dot fr}

\abstract{
Documenting languages helps to prevent the extinction of endangered dialects -- many of which are otherwise expected to disappear by the end of the century. 
When documenting oral languages, unsupervised word segmentation~(UWS) from speech is a useful, yet challenging, task.
It consists in producing time-stamps for slicing utterances into smaller segments corresponding to words, being performed from phonetic transcriptions, or in the absence of these, from the output of unsupervised speech discretization models. 
These discretization models are trained using raw speech only, producing discrete speech units that can be applied for downstream (text-based) tasks. In this paper we compare five of these models: three Bayesian and two neural approaches, with regards to the exploitability of the produced units for UWS. For the UWS task, we experiment with two models, using as our target language the Mboshi~(Bantu C25), an unwritten language from Congo-Brazzaville. Additionally, we report results for Finnish, Hungarian, Romanian and Russian in equally low-resource settings, using only 4 hours of speech. Our results suggest that neural models for speech discretization are difficult to exploit in our setting, and that it might be necessary to adapt them to limit sequence length. We obtain our best UWS results by using Bayesian models that produce high quality, yet compressed, discrete representations of the input speech signal.
\\ \newline \Keywords{unsupervised word segmentation, speech discretization, acoustic unit discovery, low-resource settings}}

\begin{document}

\maketitleabstract

\input{sections/1_introduction}

\input{sections/2_audmodels}

\input{sections/3_uwsmodels}

\input{sections/4_experiments}

\input{sections/5_discussion}

\section{Bibliographical References}\label{reference}
\bibliographystyle{lrec2022-bib}
\bibliography{lrec2022-example}

\section{Language Resource References}
\label{lr:ref}
\bibliographystylelanguageresource{lrec2022-bib}
\bibliographylanguageresource{languageresource}

\end{document}

%% file: sections/1_introduction.tex
\section{Introduction}

Popular models for speech processing still rely on the availability of considerable amounts of speech data and their transcriptions, which reduces model applicability to a limited subset of languages considered \textit{high-resource}. This excludes a considerable number of \textit{low-resource} languages, including many from oral tradition.
Besides, learning supervised representations from speech differs from the unsupervised way infants learn language, hinting that it should be possible to develop more data-efficient speech processing models.

Recent efforts for \textit{zero-resource} processing~\cite{glass2012,jansen2013JHU,versteegh2016zero,zrc2017,Dunbar2019,dunbar2020zero} focus on building speech systems using limited amounts of data (hence \textit{zero resource}), and without textual or linguistic resources, for increasingly challenging tasks such as acoustic or lexical unit discovery. Such zero resource approaches also stimulated interest for computational language documentation~\cite{besacier2006towards,duong2016attentional,Godard2018,bird2021sparse} and computational language acquisition~\cite{dupoux2016}.

In this paper we address the challenging task of unsupervised word segmentation~(UWS) from speech. This task 
consists of 
outputting time-stamps delimiting stretches of speech, associated with class labels corresponding to word hypotheses, without access to any supervision. We build on the work presented in~\newcite{Godard2018}: they proposed a cascaded model for UWS that first generates a discrete sequence from the speech signal using the model from~\newcite{ondel2016variational}, and then segments the discrete sequence into words using a Bayesian~\cite{goldwater2007nonparametric} or a neural~\cite{boito2017unwritten} approach.
Since then, much progress has been made in automatic speech discretization: efficient Bayesian models for acoustic unit discovery~(AUD) emerged~\cite{ondel2019shmm,yusuf2020hierarchical}, and self-supervised models based on neural networks -- typically made of an auto-encoder structure with a discretization layer -- were also introduced~\cite{NIPS2017_7a98af17,baevski2019vq,chorowski2019unsupervised}.

Therefore, in this work we revise and extend \newcite{Godard2018} by empirically investigating the \textit{exploitability} of five recent approaches for speech discretization for the UWS task in a rather low-resource scenario, using approximately 4 hours of speech~(roughly 5k sentences).
More precisely,  we train three Bayesian speech discretization models~(\textit{HMM}~\cite{ondel2016variational}, \textit{SHMM}~\cite{ondel2019shmm} and \textit{H-SHMM}~\cite{yusuf2020hierarchical}), and two neural models~(\textit{VQ-VAE}~\cite{NIPS2017_7a98af17} and \textit{vq-wav2vec}~\cite{baevski2019vq}). 
We extract discrete speech units from them using only 4 hours of speech, and we perform UWS from the  sequences produced. 
Our pipeline targets the Mboshi language~(Bantu C25), an unwritten language from Congo-Brazzaville. Additionally, we perform experiments in equal data settings for Finnish, Hungarian, Romanian and Russian. This allows us to assess the language-related impact in our UWS pipeline.

Our experiments show that neural models for speech discretization are difficult to exploit for UWS, as they output very long sequences. In contrast to that, the Bayesian speech discretization approaches from~\newcite{ondel2019shmm} and \newcite{yusuf2020hierarchical} are robust and generalizable, producing high quality, yet compressed, discrete speech sequences from the input utterances in all languages. 
We obtain our best results by using these sequences for training the neural UWS model from~\newcite{boito2017unwritten}.

This paper is organized as follows. Section~\ref{sec:related} presents related work, and Section~\ref{sec:aud} details the speech discretization models we experiment with. Section~\ref{sec:expsetup} presents our experimental setup, and Section~\ref{sec:exp} our experiments. Section~\ref{sec:discuss} concludes our work.

\section{Related Work}\label{sec:related}

The work presented here revises the UWS model from speech in low-resource settings presented in \newcite{Godard2018}. \newcite{Boito2019} complemented that work by tackling different neural models for bilingual UWS, but they did not address the discretization portion of the pipeline, working directly from manual phonetic transcriptions. In \newcite{kamper2020towards}, the authors propose constraining the VQ-VAE model in order to generate a more exploitable output representation for direct application to the UWS task in English. Different from that, in this work we focus on providing an empirical comparison of recent discretization approaches, extending \newcite{Godard2018} and providing results in low-resource settings, and in five different languages.

This work falls into the category of computational language documentation approaches. Recent works in this field include the use of aligned translation for improving transcription quality~\cite{anastasopoulos2018leveraging}, and for obtaining bilingually grounded UWS~\cite{duong2016attentional,boito2017unwritten}. We find 
pipelines for obtaining manual~\cite{foley2018building} and automatic~\cite{michaud2018integrating} transcriptions, and for aligning transcription and audio~\cite{strunk2014untrained}. Other examples are methods for low-resource segmentation~\cite{lignos2010recession,Goldwater09bayesian}, and for lexical unit discovery without textual resources~\cite{bartels2016toward}. Finally, direct speech-to-speech~\cite{tjandra2019speech} and speech-to-text~\cite{besacier2006towards,berard2016listen} architectures could be an option for the lack of transcription, but it remains to be seen how exploitable these architectures can be in low-resource settings.

Lastly, we highlight that recent models based on self-supervised learning~\cite{schneider2019wav2vec,baevski2019effectiveness,wang2020unsupervised,liu2020mockingjay,baevski2020wav2vec,hsu2021hubert} provide an interesting novel option for reducing the amount of labeled data needed in downstream tasks such as automatic speech recognition and speech translation. 
In this work we experiment with the vq-wav2vec model, a predecessor of the popular wav2vec~2.0~\cite{baevski2020wav2vec}.
We however, do not extend our investigation to the latter, or to models such as HuBERT~\cite{hsu2021hubert}. This is because, while these models do produce a certain discretization of the speech~(for wav2vec~2.0 via quantization module, for HuBERT via clustering of MFCC features), we judge this discretization to be insufficiently exploitable for downstream text-based approaches due to their excessive length.\footnote{For instance, wav2vec~2.0 trains on a joint \textit{diversity} loss for inciting the use of its discrete units. Their large codebook of $G=8;V=8$ results in an upper-bound of $8^8$ units.} We do, however, find promising the integration of self-supervised speech features into Bayesian AUD models as in~\newcite{Ondel2022}.

%% file: sections/2_audmodels.tex
\section{Unsupervised Speech Discretization Models}\label{sec:aud}

Speech discretization consists in labeling the speech signal into discrete speech units, which can correspond or not to the language phonetic inventory. This problem can be formulated as the learning of a set of $U$ discrete units with embeddings $\Matrix{H} = \{\Matrix{\eta}^1, \dots, \Matrix{\eta}^U\}$ from a sequence of untranscribed acoustic features $\Matrix{X}= [\Matrix{x}_1, \dots, \Matrix{x}_N]$, as well as the assignment of frame to unit $\Matrix{z} = [z_1, \dots, z_N]$. Depending on the approach, neural~(Section~\ref{sec:aud:neural}) or Bayesian~(Section~\ref{sec:aud:bayesian}), the assumptions and the inference regarding these three quantities will differ.

\subsection{Neural (VQ-based) models}\label{sec:aud:neural}

\paragraph{VQ-VAE.}
It comprises an encoder, a decoder, and a set of unit-specific embeddings $\Matrix{H}$. The encoder is a neural network that transforms the data into a continuous latent representation $\Matrix{V} = (\Matrix{v}_1, \dots, \Matrix{v}_N)$. Each frame is then assigned to the closest embedding in the Euclidean sense~(Equation~\ref{eq:vqvae1}). The decoder transforms the sequence of quantized vectors into parameters of the conditional log-likelihood of the data $p(\Matrix{x}_n | \Matrix{z})$, and the network is trained to maximize this likelihood. Since the quantization step is not differentiable, the encoder is trained with a straight through estimator~\cite{bengio2013estimating}. In addition, a pair of $\ell_2$ losses are used to minimize the quantization error, and the overall objective function that is maximized is presented in Equation~\ref{eq:vqvae2}, where $\sg[\cdot]$ is the stop-gradient operator. We define the likelihood $p(\Matrix{x}_n | z_n) = \Gaussian(\Matrix{x}_n; \Matrix{\mu}(\Matrix{\eta}^{z_n}), \Matrix{I})$. Under this assumption, the log-likelihood reduces to the mean-squared error $||\Matrix{x}_n - \Matrix{\mu}(\Matrix{\eta}^{z_n})||_2^2$.
\resizebox{0.9\linewidth}{!}{
\begin{minipage}{\linewidth}
\begin{align}
    z_n = \argmin_u ||\Matrix{v}_n - \Matrix{\eta}^u||_2.
    \label{eq:vqvae1}
\end{align}
\end{minipage}}
\resizebox{0.9\linewidth}{!}{
\begin{minipage}{\linewidth}
\begin{align}
    \mathcal{L} = \frac{1}{N} \sum_{n=1}^N \Bigl ( & \ln p(\Matrix{x}_n | z_n) - k_1||\sg[\Matrix{\eta}^{z_n}] - \Matrix{v}_n||_2^2 \nonumber \\
    &- k_2||\Matrix{\eta}^{z_n} - \sg[\Matrix{v}_n]||_2^2 \Bigr ),
    \label{eq:vqvae2}
\end{align}
\end{minipage}}

\paragraph{vq-wav2vec.}
This model is composed of an encoder~($f : \Matrix{X} \xrightarrow{} \Matrix{Z}$), a quantizer~($q : \Matrix{Z} \xrightarrow{} \Matrix{\hat{Z}}$) and an aggregator~($g : \Matrix{\hat{Z}} \xrightarrow{} \Matrix{C}$). 
The encoder is a CNN which maps the raw speech input $\Matrix{X}$ into the dense feature representation $\Matrix{Z}$. From this representation, the quantizer produces discrete labels $\Matrix{{\hat{Z}}}$ from a fixed-size codebook $\Matrix{e} \in \mathbb{R}^{V\times d}$ with $V$ representations of size $d$. 
Since replacing an encoder feature vector $\Matrix{z}_i$ by a single entry in the codebook makes the method prone to model collapse, 
the authors independently quantize partitions of each feature vector by 
creating multiple \textit{groups} $G$, arranging the feature vector into a matrix $\Matrix{z'} \in \mathbb{R}^{G\times(d/G)}$. Considering each row as an integer index, the 
full feature vector is represented by the indices $\Matrix{i} \in [V]^G$, with $V$ being the possible number of \textit{variables} for a given group, 
and each element $\Matrix{i}_j$ corresponding to a fixed codebook vector ($j \in |G|$).
For each of the $G$ groups, the quantization is performed by using Gumbel-Softmax~\cite{jang2017categorical} or online k-means clustering.
Finally, the aggregator combines multiple quantized feature vector time-steps into a new representation $\Matrix{c}_i$ for each time step $i$. The model is trained 
to distinguish a sample $k$ steps in the future $\Matrix{\hat{z}}_{i+k}$ 
from \textit{distractor} samples $\Matrix{\Tilde{z}}$ drawn from a distribution $p_n$. This is done by minimizing the contrastive loss for steps $k=\{1,\dots,K\}$ as in Equation~\ref{eq:vqwav2vec1},  where $T$ is the sequence length, $\sigma (x) = 1/(1+exp(-x))$, 
$\sigma (\Matrix{\hat{z}}_{i+k}^\intercal h_k(\Matrix{c_i}))$ is the probability of $\Matrix{\hat{z}}_{i+k}$ being the true sample, and $h_k(\Matrix{c}_i)$ is the 
step-specific affine transformation $h_k(\Matrix{c}_i) = W_k\Matrix{c}_i + b_k$. Finally, this loss is accumulated over all $k$ steps $\mathcal{L} = \sum_{k=1}^K \mathcal{L}_k$.

\resizebox{0.9\linewidth}{!}{
\begin{minipage}{\linewidth}
\begin{align}
    \mathcal{L}_k = \sum_{i=1}^{T-k} \Bigl( &\log \sigma (\Matrix{\hat{z}}_{i+k}^\intercal h_k(\Matrix{c_i})) \nonumber \\
    &+ \lambda\mathbb{E}_{\Matrix{\Tilde{z}}\sim p_n}[\log \sigma (-\Matrix{\Tilde{z}}^\intercal h_k(\Matrix{c}_i)) ] \Bigl) 
    \label{eq:vqwav2vec1}
\end{align}
\end{minipage}}

\paragraph{Training.} 
For \textbf{VQ-VAE}, the encoder has 4 Bi-LSTM layers each with output dimension 128 followed by a 16-dimensional feed-forward decoder with one hidden layer. The number of discovered units (quantization centroids) is set to 50. This setting is unusually low but it helps to reduce the length of the output sequence. We set $k_1 = 2$ and $k_2 = 4$~(Equation~\ref{eq:vqvae2}), 
and train\footnote{Implementation available at: \url{https://github.com/BUTSpeechFIT/vq-aud}} 
with Adam~\cite{adam} with an initial learning rate of $2\times10^{-3}$ which is halved whenever the loss stagnates for two training epochs.

For \textbf{vq-wav2vec}, we use the small model from \cite{baevski2019vq},\footnote{Implementation available at: \url{https://github.com/pytorch/fairseq/tree/master/examples/wav2vec}} 
but with only 64 channels, residual scale of $0.2$, and warm-up of 10k. For vocabulary we set $G=2$ and experimented with having both $V=4$, resulting in 16 units~(\textit{VQ-W2V-V16}), and $V=6$, resulting in 36 units~(\textit{VQ-W2V-V36}). Larger vocabularies resulted in excessively long sequences which could not be used for UWS.\footnote{For instance, the \texttt{dpseg} original implementation only processes sequences shorter than 350 tokens.}
We also experimented reducing the representation by using byte pair encoding~(BPE)~\cite{sennrich-etal-2016-neural}, hypothesizing  that phones were being modeled by a combination of different units. In this setting, BPE serves as a method for identifying and clustering these patterns. 
Surprisingly, we found that using BPE resulted in a decrease in UWS performance. This hints that this model might not be very consistent during its labeling process.

\subsection{Bayesian Generative Models}\label{sec:aud:bayesian}
For generative models, each acoustic unit embedding $\Matrix{\eta}_i$ represents the parameters of a probability distribution $p(\Matrix{x}_n | \Matrix{\eta}_{z_n}, z_n )$ with latent variables $\Matrix{z}$. Discovering the units amounts to estimating the posterior distribution over the embeddings $\Matrix{H}$ and the assignment variables $\Matrix{z}$ given by:

\resizebox{0.9\linewidth}{!}{
\begin{minipage}{\linewidth}
\begin{align}
    p(\Matrix{z}, \Matrix{H} | \Matrix{X}) &\propto p(\Matrix{X} | \Matrix{z}, \Matrix{H}) p(\Matrix{z} | \Matrix{H}) \prod_{u=1}^{U}p(\Matrix{\eta}^u). 
    \label{eq:bayes}
\end{align}
\end{minipage}
}

From this, we describe three different approaches.

\paragraph{HMM.} In this model each unit is a 3-state left-to-right HMM with parameters $\Matrix{\eta}^{i}$. Altogether, the set of units forms a large HMM analog to a ``phone-loop'' recognition model. This model, described in \newcite{ondel2016variational}, serves as the backbone for the two subsequent models.

\paragraph{SHMM.} The prior $p(\Matrix{\eta})$ in Equation~\ref{eq:bayes} is the probability that a sound, represented by an HMM with parameters $\Matrix{\eta}$, is an acoustic unit. For the former model, it is defined as a combination of exponential family distributions forming a prior conjugate to the likelihood. While mathematically convenient, this prior does not incorporate any knowledge about phones, i.e. it considers all possible sounds as potential acoustic units.
In \newcite{ondel2019shmm}, they propose to remedy this shortcoming by defining the parameters of each unit $u$ as in Equation~\ref{eq:shmm}, where $\Matrix{e}^u$ is a low-dimensional unit embedding, $\Matrix{W}$ and $\Matrix{b}$ are the parameters of the \emph{phonetic subspace}, and the function $f(\cdot)$  ensures that the resulting vector $\Matrix{\eta}^u$ dwells in the HMM parameter space. 
The subspace, defined by $\Matrix{W}$ and $\Matrix{b}$, is estimated from several labeled source languages. 
The prior $p(\Matrix{\eta})$ is defined over the low-dimensional embeddings $p(\Matrix{e})$ rather than $\Matrix{\eta}$ directly, therefore constraining the search of units in the relevant region of the parameter space. This model is denoted as the Subspace HMM (SHMM). 
\resizebox{0.9\linewidth}{!}{
\begin{minipage}{\linewidth}
\begin{align}
    \Matrix{\eta}^u =f( \Matrix{W} \cdot \Matrix{e}^u + \Matrix{b})
    \label{eq:shmm}
\end{align}
\end{minipage}
}

\paragraph{H-SHMM.} While the SHMM significantly improves results 
over the HMM, it also suffers from an unrealistic assumption: it assumes that the phonetic subspace is the same for all languages. \newcite{yusuf2020hierarchical} relax this assumption by proposing to adapt the subspace for each target language while learning the acoustic units. Formally, for a given language $\lambda$, the subspace and the acoustic units' parameters are constructed as in Equation~\ref{eq:W}-\ref{eq:eta}, where the matrices $\Matrix{M}_0, \dots, \Matrix{M}_K$ and vectors $\Matrix{m}_0, \dots, \Matrix{m}_K$ represent some ``template'' phonetic subspace linearly combined by a language embedding $\Matrix{\alpha}^{\lambda} = [\alpha^\lambda_1, \alpha^\lambda_2, \dots, \alpha^\lambda_K]^\top$. The matrices $\Matrix{M}_i$ and the vectors $\Matrix{m}_i$ are estimated from labeled languages -- from 
multilingual transcribed speech dataset for instance. 
The acoustic units' low-dimensional embeddings $\{ \Matrix{e}_i \}$ and the language embedding $\Matrix{\alpha}$ are learned on the target~(unlabeled) speech data.
We refer to this model as the Hierarchical SHMM~(H-SHMM).

\resizebox{0.9\linewidth}{!}{
\begin{minipage}{\linewidth}
\begin{align}
    \Matrix{W}^\lambda &= \Matrix{M}_0 + \sum_{k=1}^K \alpha^\lambda_k \Matrix{M_k} \label{eq:W} \\
    \Matrix{b}^\lambda &= \Matrix{m}_0 + \sum_{k=1}^K \alpha^\lambda_k \Matrix{m_k} \label{eq:b} \\
    \Matrix{\eta}^{\lambda,u} &= f(\Matrix{W}^\lambda \cdot \Matrix{e}^{\lambda, u} + \Matrix{b}^\lambda)
    \label{eq:eta}
\end{align} 
\end{minipage}
}

\paragraph{Inference.} For the three generative models, the posterior distribution is intractable and cannot be estimated. Instead, one seeks an approximate posterior $q(\{ \Matrix{\eta}_i \} , \Matrix{z}) = q(\{ \Matrix{\eta}_i \})q(\Matrix{z})$ that maximizes the variational lower-bound $\mathcal{L}[q]$. 
Concerning the estimation of $q(\Matrix{z})$, the \textit{expectation} step is identical for all models and is achieved with a modified \textit{forward-backward} algorithm described in \newcite{ondel2016variational}. Estimation of $q(\Matrix{\eta})$, the \textit{maximization} step, is model-specific and is described in \newcite{ondel2016variational} for the HMM, in \newcite{ondel2019shmm} for SHMM models, and in \newcite{yusuf2020hierarchical} for the H-SHMM model.
Finally, the output of each 
system is obtained from a modified Viterbi algorithm that uses the expectation of the log-likelihoods with respect to $q(\{ \Matrix{\eta}_i \})$, instead of point estimates.

\paragraph{Training.} The models 
are trained  with 4 Gaussians per HMM state and using 100 for the Dirichlet process' truncation parameter. SHMM and H-SHMM use an embedding size of 100, and H-SHMM models have a 6-dimensional language embedding. For the methods that use subspaces estimation (SHMM and H-SHMM), this estimation uses the following languages:
French, German, Spanish, Polish from the Globalphone corpus~\citelanguageresource{schultz2013globalphone}, as well as Amharic~\citelanguageresource{Abate2005amharic}, Swahili~\citelanguageresource{gelas2012swahili} and Wolof~\citelanguageresource{gauthier2016wolof} from the ALFFA project~\cite{besacier2015speech}. We use 2-3 hours subsets of each, for a total of roughly 19 hours.

%% file: sections/3_uwsmodels.tex
\section{Experimental Setup}\label{sec:expsetup}

\input{tables/datasets_stats}
\input{tables/units_stats}

From the discrete speech units produced by the presented speech discretization models, we produce segmentation in the symbolic domain by using two UWS models.
A final speech segmentation is then inferred using the units' time-stamps and 
evaluated by using the \textit{Zero-Resource Challenge} 
2017 evaluation suite, track~2~\cite{zrc2017}\footnote{Resources are available at \url{http://zerospeech.com/2017}}. We now detail the UWS models used in this work, which are trained with the same parameters from~\newcite{Godard2018}. We also detail the datasets and the post-processing for the discrete speech discrete units. 

\paragraph{Bayesian UWS approach (monolingual).} Non-parametric Bayesian models~\cite{goldwater2007nonparametric,johnson2009improving} are statistical approaches for UWS and morphological analysis, known to be robust in low-resource settings~\cite{godard2016preliminary}. In these models, words are generated by a unigram or bigram model over an infinite inventory, through the use of a Dirichlet process. In this work, we use the unigram model from \textit{dpseg}~\cite{Goldwater09bayesian}\footnote{Implementation available at \url{http://homepages.inf.ed.ac.uk/sgwater/resources.html}}, which was shown to be superior to the bigram model in low-resource settings~\cite{godard2019unsupervised}.

\paragraph{Neural UWS approach (bilingual).} We follow the bilingual pipeline from~\newcite{Godard2018}. 
The discrete speech units and their sentence-level translations are fed to an attention-based neural machine translation system that produces soft-alignment probability matrices between source and target sequences. For each sentence pair, its matrix is used for clustering together~(segmenting) neighboring phones whose alignment distribution peaks at the same source word.
Examples of these matrices are provided in Figure~\ref{fig:heatmaps}.  We refer to this model as \textit{neural}.

\input{figures/neural_heatmaps}

\noindent\textbf{Datasets.} We use the Mboshi-French parallel corpus~(MB-FR)~\citelanguageresource{godard2017very}, which is a 5,130 sentence corpus from the language documentation process of Mboshi~(Bantu C25), an oral language spoken in Congo-Brazzaville. We also report results using an extract from the MaSS corpus~\citelanguageresource{boito2020mass}, a multilingual speech-to-speech and speech-to-text dataset. 
We use the down-sampling from~\newcite{boito2020investigating}, which results in 5,324 aligned sentences. We exclude French and Spanish, as these languages are present in the subspace prior from SHMM and H-SHMM models, and we exclude English as it was used as to tune the hyperparameters of the subspace models and the VQ-VAE.  We also exclude Basque, as the  sequences produced were too long for UWS training. The final set of languages is: Finnish~(FI), Hungarian~(HU), Romanian~(RO) and Russian~(RU).
In all cases, the French~(FR) translations are used as supervision for the neural UWS approach.
Statistics are presented in Table~\ref{tab:corpusstats}.


\paragraph{Discrete Speech Units Post-processing.} 
We experiment with reducing the representation by removing units predicted in silence windows. For this, we use the gold references' silence annotations. Removing these allow us to focus the investigation on the quality of the units generated in \textit{relevant} portions of the speech. We see in Table~\ref{tab:unitsstats} that removing windows that we \textit{know} correspond to silence considerably reduces the number of units generated by all models. Before UWS evaluation, the silence windows are reintroduced to ensure that their segmentation boundaries are taken into account. This approach is justified because a silence detector is an inexpensive resource to obtain. For instance, popular software such as Praat~\cite{boersma2006praat} are able to handle this task in any language. 
Figure~\ref{fig:unitssegmentation} exemplifies the discrete speech units discovered by the models before applying this post-processing.

%% file: tables/datasets_stats.tex
\begin{table}
\scriptsize
\centering
\resizebox{\columnwidth}{!}{
\begin{tabular}{clcccc}\hline
                                &              & \textbf{\#Types} & \textbf{\#Tokens} & \textbf{\begin{tabular}[c]{@{}c@{}}Avg Token \\ Length\end{tabular}} & \textbf{\begin{tabular}[c]{@{}c@{}}Avg \#Tokens \\ per Sentence\end{tabular}} \\\hline
\multirow{2}{*}{\textbf{MB-FR}} & \textbf{MB*} & 6,633             & 30,556            & 4.2                                                                  & 6.0                                                                           \\
                                & \textbf{FR}  & 5,162             & 42,715            & 4.4                                                                  & 8.3                                                                           \\\hline
\multirow{5}{*}{\textbf{MaSS}}  & \textbf{FI*} & 12,088            & 70,226            & 6.0                                                                  & 13.2                                                                          \\
                                & \textbf{HU*} & 12,993            & 69,755            & 5.9                                                                  & 13.1                                                                          \\
                                & \textbf{RO*} & 6,795             & 84,613            & 4.5                                                                  & 15.9                                                                          \\
                                & \textbf{RU*} & 10,624            & 67,176            & 6.2                                                                  & 12.6                                                                          \\
                                & \textbf{FR}  & 7,226             & 94,527            & 4.1                                                                  & 17.8                                                                         \\\hline
\end{tabular}}
\caption{Statistics for the datasets, computed over the text (FR), or over the phonetic representation (*).}
\label{tab:corpusstats}
\end{table}

%% file: tables/units_stats.tex
\begin{table}
\centering
\scriptsize
\resizebox{\columnwidth}{!}{
\begin{tabular}{llccc}\hline
\textbf{}                     &                                                                               & \textbf{HMM}    & \textbf{SHMM}     & \textbf{H-SHMM}   \\\hline
\multirow{3}{*}{\rotatebox[origin=c]{90}{\textbf{RAW}}} & \textbf{\# Units}                                                             & 77 (+9)         & 76 (+8)           & 49 (-19)          \\
                              & \textbf{\begin{tabular}[c]{@{}l@{}}Avg \#Units \\ per sequence\end{tabular}} & 27.5 (+8.7)     & 24.0 (+5.2)       & 21.7 (+2.9)       \\
                              & \textbf{Max Length}                                                           & 68 (+17)        & 69 (+18)          & 63 (+12)          \\\hline
\multirow{3}{*}{\rotatebox[origin=c]{90}{\textbf{+SIL}}}  & \textbf{\# Units}                                                             & 75 (+7)         & 75 (+7)           & 47 (-21)          \\
                              & \textbf{\begin{tabular}[c]{@{}l@{}}Avg \#units \\ per sequence\end{tabular}} & 20.9 (+2.1)     & 19.9 (+1.1)       & 19.4 (+0.6)       \\
                              & \textbf{Max Length}                                                           & 69 (+18)        & 62 (+11)          & 60 (+9)           \\\hline
\textbf{}                     & \textbf{}                                                                     & \textbf{VQ-VAE} & \textbf{VQ-W2V-16} & \textbf{VQ-W2V-36} \\\hline
\multirow{3}{*}{\rotatebox[origin=c]{90}{\textbf{RAW}}}  & \textbf{\# Units}                                                             & 50 (-18)        & 16 (-52)          & 36 (-32)          \\
                              & \textbf{\begin{tabular}[c]{@{}l@{}}Avg \#units \\ per sequence\end{tabular}} & 65.2 (+46.4)    & 81.7 (+62.9)      & 111.0 (+92.2)     \\
                              & \textbf{Max Length}                                                           & 217 (+166)      & 289 (+238)        & 361 (+310)        \\\hline
\multirow{3}{*}{\rotatebox[origin=c]{90}{\textbf{+SIL}}} & \textbf{\# Units}                                                             & 50 (-18)        & 16 (-52)          & 36 (-32)          \\
                              & \textbf{\begin{tabular}[c]{@{}l@{}}Avg \#units \\ per sequence\end{tabular}} & 43.4 (+24.6)    & 52.6 (+33.8)      & 76.2 (+57.4)      \\
                              & \textbf{Max Length}                                                           & 143 (+92)       & 229 (+178)        & 271 (+220)     \\\hline  
\end{tabular}}
\caption{Statistics for the discrete speech units produced for the Mboshi, with the difference between the produced and reference representation between parentheses. RAW is the original output from speech discretization models, +SIL is the result after silence post-processing. Other languages follow the same trend.}
\label{tab:unitsstats}
\end{table}

%% file: figures/neural_heatmaps.tex
\begin{figure}
    \centering
    \resizebox{\columnwidth}{!}{
    \includegraphics{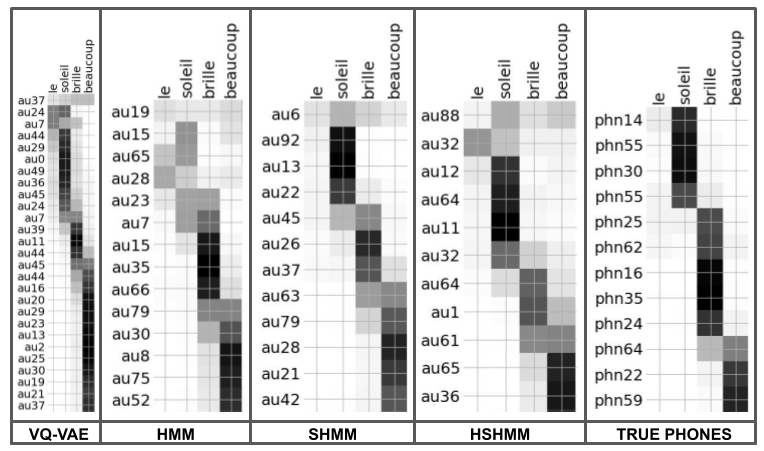}}
 \caption{Heatmaps for the soft-alignment probability matrices generated by the neural UWS models (bilingual) trained on different discrete speech units, for the same French-Mboshi sentence. The darker the square, the higher the pair probability. The rows present the automatically generated units from the different discretization models, informed in the bottom.}   
    \label{fig:heatmaps}
\end{figure}

%% file: sections/4_experiments.tex
\section{Experiments}\label{sec:exp}

\input{tables/mboshi_uws_results}
\input{figures/units_example_small}

We first present our results for the MB-FR dataset, the language which corresponds to the true low-resource scenario that we are interested in. Table~\ref{tab:mboshi_uws_results} presents UWS Boundary F-scores for UWS models~(dpseg and neural) trained using different discrete speech units for the MB-FR dataset. We include results for both the direct output~(RAW) and the post-processed version~(+SIL). The RAW VQ-W2V-V36 is not included as its output sequences 
were excessively large for training our UWS models~(Table~\ref{tab:unitsstats}).

We observe that in all cases, post-processing the discrete speech units with the silence information~(+SIL) creates \textit{easier} representations for the UWS task. We believe this is due to the considerable reduction in average length of the sequences~(Table~\ref{tab:unitsstats}). For Bayesian models, we also observe a reduction in the number of units, meaning that some units were modelling silence windows, even though these models already produce an independent token for 
silence, which we remove before UWS training.

Looking at the results for UWS models trained using the output of VQ-based models~(rows 4-6), we see that the best segmentation result is achieved using the one with the smallest average sequence length~(VQ-VAE). In general, we believe that all VQ-based models under-perform due to the excessively long sequences produced, which are challenging for UWS. Figure~\ref{fig:unitssegmentation} illustrates this difference in representation length, by presenting the discrete speech units produced by Bayesian and neural models 
for a given utterance: the latter produce considerably more units.

Overall, we find that UWS models trained using the discrete speech units from Bayesian models produce better segmentation, with models trained with SHMM and H-SHMM presenting the best results. In \newcite{yusuf2020hierarchical} both systems showed competitive results for the AUD task. 
A noticeable difference between  these two models is the compression level: H-SHMM uses 27 fewer units than SHMM. Regarding type retrieval, the models scored 12.1\%~(SHMM), 10.7\%~(H-SHMM), and 31\%~(topline).
We also find that SHMM models produced more types and fewer tokens, reaching a higher Type-Token Ratio~(0.63) compared to H-SHMM~(0.55). 

\input{tables/mass_uws_results}

Focusing on the generalization of the presented speech discretization models, we trained our models using four languages from the MaSS dataset. We observed that due to the considerably larger average length of the sentences~(Table~\ref{tab:corpusstats}), the VQ-based models produced sequences which we were unable to directly apply to UWS training. This again highlights that these models need some constraining, or post-processing, in order to be directly exploitable for UWS.
Focusing on the Bayesian models, which performed the best for generating exploitable discrete speech units for UWS in low-resource settings, 
Table~\ref{tab:massuwsresults} present UWS results. 
We omit results for RAW, as we observe the same trend from Table~\ref{tab:mboshi_uws_results}. Looking at the results for the four languages, 
we again observe competitive results for SHMM and H-SHMM models, illustrating that these approaches generalize well to different languages.

Comparing the UWS results present in Table~\ref{tab:mboshi_uws_results}~(Mboshi) and Table~\ref{tab:massuwsresults}~(languages from MaSS),
we notice overall lower results for the languages from the MaSS dataset~(best result:~59.6) compared to Mboshi~(best result:~64.7). We believe this is due to the MaSS data 
coming from read text, in which the utterances correspond to verses that 
are consistently longer than sentences~(Table~\ref{tab:corpusstats}). This results in a more challenging setting for UWS and explains the lower results.
Lastly, 
our results over five languages
show that the neural UWS model produces better segmentation results from discrete speech units than dpseg, which in turn performs the best with the true phones~(topline). This confirms the trend observed by~\cite{Godard2018}. 
The neural UWS models have the advantage of their word-level aligned translations for grounding the segmentation process, which might be attenuating the difficulty of the task in this noisier scenario, with longer sequences and more units.
Moreover, a benefit of these models is the potentially exploitable bilingual alignment discovered during training. \newcite{Boito2019} used these alignments for filtering the generated vocabulary, increasing type retrieval. 

%% file: tables/mboshi_uws_results.tex
\begin{table}
\centering
\scriptsize
\resizebox{\columnwidth}{!}{
\begin{tabular}{llcc|cc}\hline
         &               & \multicolumn{2}{c|}{\textbf{\textit{dpseg}}} & \multicolumn{2}{c}{\textbf{\textit{neural}}} \\\hline
 &                       & \textbf{RAW}              & \textbf{+SIL}            & \textbf{RAW}              & \textbf{+SIL}             \\
\textbf{1}&\textbf{HMM}           & 32.4             & 59.9            & 35.1             & 61.2             \\
\textbf{2}&\textbf{SHMM}           & 43.7             & \textbf{61.4}            & 41.4             & \textbf{64.7}             \\
\textbf{3}&\textbf{H-SHMM}         & \textbf{45.3}             & \textbf{61.4}            & \textbf{44.8}             & 63.9             \\
\textbf{4}&\textbf{VQ-VAE}         & 39.0             & 52.7            & 32.1             & 60.1             \\
\textbf{5}&\textbf{VQ-W2V-V16} & 37.4             & 52.2            & 32.0             & 50.6             \\
\textbf{6}&\textbf{VQ-W2V-V36} & -                & 48.0            & -                & 49.8             \\\hline
\textbf{7}&\textbf{True Phones}    & -                & \textbf{77.1}            & -                & 74.5            \\\hline
\end{tabular}}
\caption{UWS Boundary F-scores for the MB-FR dataset.}
\label{tab:mboshi_uws_results}
\end{table}

%% file: figures/units_example_small.tex
\begin{figure}
    \centering
    \begin{subfigure}[b]{7.9cm}
         \centering
         \includegraphics[width=\textwidth]{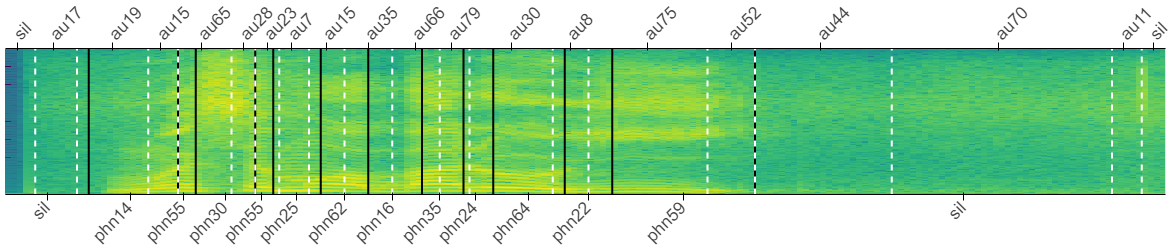}
         \caption{HMM}
         \label{fig:hmm_seg}
     \end{subfigure}
    \centering
    \begin{subfigure}[b]{7.9cm}
         \centering
         \includegraphics[width=\textwidth]{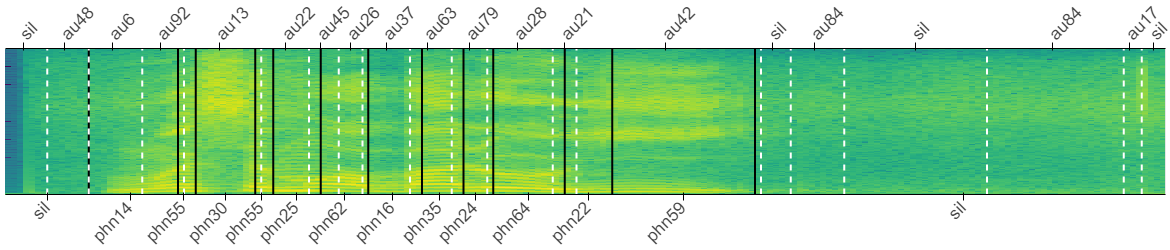}
         \caption{SHMM}
         \label{fig:shmm_seg}
     \end{subfigure}
    \centering
    \begin{subfigure}[b]{7.9cm}
         \centering
         \includegraphics[width=\textwidth]{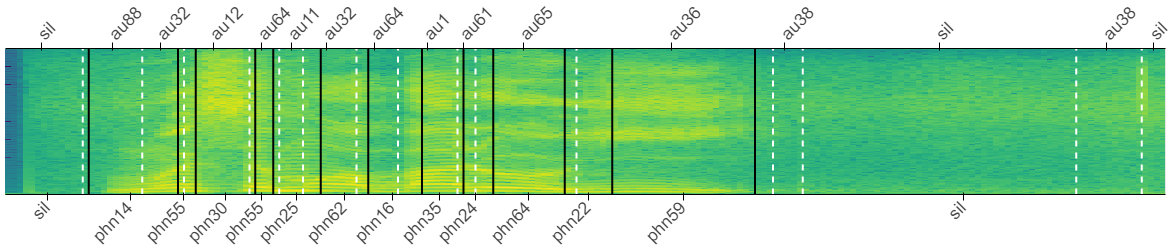}
         \caption{H-SHMM}
         \label{fig:hshmm_seg}
     \end{subfigure}
     \centering
     \begin{subfigure}[b]{7.9cm}
         \centering
         \includegraphics[width=\textwidth]{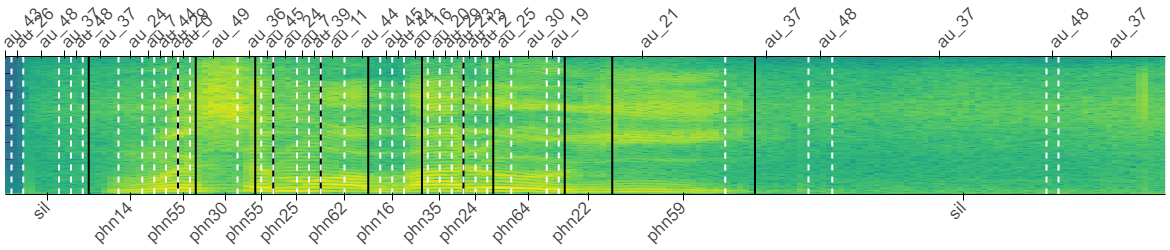}
         \caption{VQ-VAE}
         \label{fig:vqvae_seg}
     \end{subfigure}
     \centering
     \begin{subfigure}[b]{7.9cm}
         \centering
         \includegraphics[width=\textwidth]{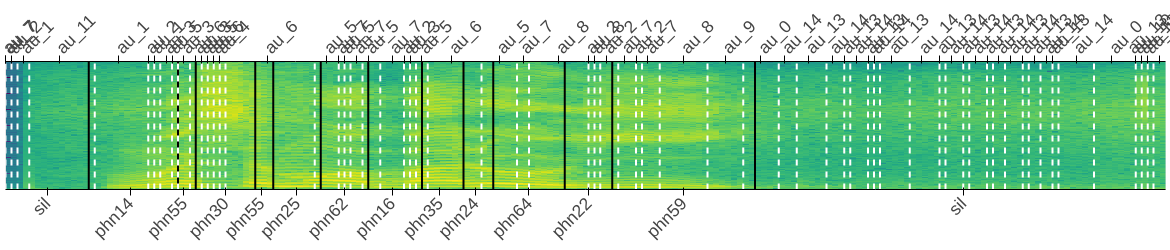}
         \caption{VQ-W2V-V16}
         \label{fig:vq2}
     \end{subfigure}
     \centering
     \begin{subfigure}[b]{7.9cm}
         \centering
         \includegraphics[width=\textwidth]{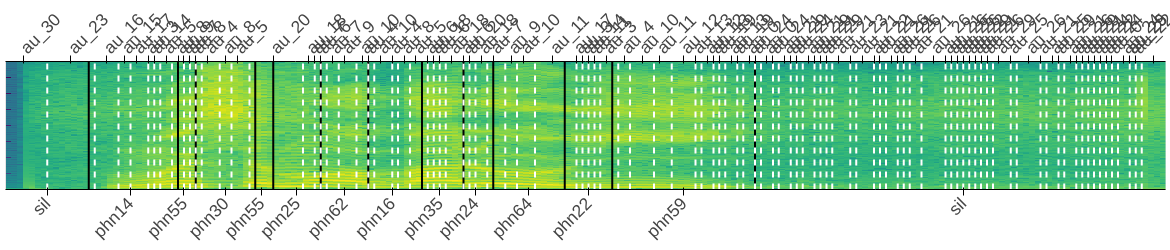}
         \caption{VQ-W2V-V36}
         \label{fig:vq3}
     \end{subfigure}
    \caption{Speech discrete units produced by the five models for the same Mboshi sentence. Black lines denote the true boundaries, 
    while dashed white lines denote the discovered units boundaries. For each example, discrete speech units~(top) and reference~(bottom).}
    \label{fig:unitssegmentation}
\end{figure}

%% file: tables/mass_uws_results.tex
\begin{table}
\centering
\scriptsize
\resizebox{\columnwidth}{!}{
\begin{tabular}{lcccc|cccc} \hline
                     & \multicolumn{4}{c|}{\textbf{\textit{dpseg}}}                    & \multicolumn{4}{c}{\textbf{\textit{neural}}}                   \\\hline
                     & \textbf{FI} & \textbf{HU} & \textbf{RO} & \textbf{RU} & \textbf{FI} & \textbf{HU} & \textbf{RO} & \textbf{RU} \\
\textbf{HMM}         & 45.6        & 49.9        & 53.5        & 47.1        & 53.4        & 51.2        & 56.6        & 54.9        \\
\textbf{SHMM}        & 49.0        & 52.3        & 53.5        & 50.5        & 56.0        & \textbf{53.9}        & 57.7        & \textbf{57.7}        \\
\textbf{H-SHMM}      & 50.5        & 52.9        & 58.0        & 52.9        & \textbf{56.1}        & 53.3        & \textbf{59.6}        & 56.0        \\\hline
\textbf{True Phones} & \underline{87.1}        & \underline{83.3}        & \underline{88.0}        & \underline{85.9}        & 68.4        & 63.4        & 75.7        & 68.4     \\\hline  
\end{tabular}}
\caption{UWS Boundary F-scores for the MaSS dataset using Bayesian models (+SIL only). Best UWS results from speech discrete units~(\textbf{bold}) and from true phones~(\underline{underlined}) are highlighted.}
\label{tab:massuwsresults}
\end{table}

%% file: sections/5_discussion.tex
\section{Conclusion}\label{sec:discuss}

In this paper we compared five methods for speech discretization, two neural models~(VQ-VAE, VQ-WAV2VEC), and three Bayesian approaches~(HMM, SHMM, H-SHMM), with respect to their performance serving as direct input to the task of unsupervised word segmentation~(UWS) in low-resource settings.
Our motivation for such a study lies in the need of processing oral and low-resource languages, for which obtaining transcriptions is a known bottleneck~\cite{brinckmann2009transcription}. 

In our UWS setting, and using five different languages (Finnish, Hungarian, Mboshi, Romanian and Russian), we find that VQ-based methods are not a good fit for our pipeline, as they output very long and inconsistent sequences, which are difficult to treat. This was also recently observed in~\newcite{kamper2020towards}.

In contrast to that, the Bayesian SHMM and H-SHMM models perform the best, as they produced concise yet highly exploitable representations from just few hours of speech. We believe this difference in performance is due to HMM-based models explicitly performing acoustic unit discovery. 
This means the discretization produced by them aims not only to summarize the speech signal, but to closely match 
the language's phonology. Moreover, the subspace estimation performed by both SHMM and H-SHMM, might also play a significant role. This is because these models are able to learn from an additional 19~hours of data in different languages. The other models~(HMM and VQ-based models) do not have access to any form of pretraining or prior.

Finally, comparing the neural and Bayesian UWS approaches, 
we notice that the neural model is competitive in the \textit{noisier} setting, reaching better UWS boundary scores working with the output of speech discretization models. The Bayesian model is however better at segmenting true phones~(topline scenario). Concluding, this work updates \newcite{Godard2018} by using more recent speech discretization models, and presenting better UWS results for Mboshi.